\PassOptionsToPackage{numbers,sort&compress}{natbib}
\documentclass{article}

 \usepackage[preprint]{neurips_2026}

\usepackage[utf8]{inputenc} %
\usepackage[T1]{fontenc}    %
\usepackage{hyperref}       %
\usepackage{url}            %
\usepackage{booktabs}       %
\usepackage{amsmath}       %
\usepackage{amsfonts}       %
\usepackage{nicefrac}       %
\usepackage{microtype}      %
\usepackage{xcolor}         %
\usepackage{graphicx}       %
\usepackage{subcaption}     %

\usepackage[table]{xcolor}
\usepackage{multirow}
\usepackage{booktabs}
\usepackage[table]{xcolor}

\usepackage{xcolor}
\usepackage{hyperref} 
\hypersetup{ 
     colorlinks=true,
     urlcolor=magenta, 
}

\newcommand{\ie}{\textit{i.e.}}
\newcommand{\eg}{\textit{e.g.}}

\title{Global Context Guidance for Diffusion Large Language Models}

\author{%
  {\bf Jeimin Jeon$^{1,2}$} \hspace{15mm} {\bf Bumsub Ham$^{1}$\thanks{Corresponding author}}\\
  \\
  $^{1}$Yonsei University \hspace{8mm} $^{2}$Articron Inc.  \\ \\
  \url{https://cvlab.yonsei.ac.kr/projects/GCG/}
}

\begin{document}

\maketitle

\begin{abstract}
Diffusion large language models (LLMs) have recently emerged as a promising alternative to autoregressive LLMs, enabling parallel generation through iterative denoising. Recent guidance methods improve diffusion LLMs by introducing a weak model to guide the prediction of the base model during denoising. However, the influence of the weak model during denoising remains less clear, and computing its prediction requires an additional forward pass at each denoising step. We present in this paper an empirical analysis of how diffusion LLMs use contextual information during denoising. Our analysis reveals that diffusion LLMs rely more heavily on local context while often underutilizing global context. Based on this observation, we introduce Global Context Guidance (GCG), a feature-space guidance method that enhances global-context utilization without constructing an explicit weak model. Extensive experiments on LLaDA-8B, Dream-7B, and LLaDA-1.5 across diverse benchmarks demonstrate that GCG consistently outperforms existing guidance methods, while incurring much lower inference overhead.
\end{abstract}

\section{Introduction}
Autoregressive large language models (LLMs)~\cite{grattafiori2024llama,yang2025qwen3,liu2024deepseek} have become a dominant paradigm for language generation, achieving remarkable performance across a wide range of tasks through next-token prediction. By generating text in a left-to-right manner, they model linguistic structure effectively, {showing strong in-context learning and reasoning capabilities}. However, their sequential decoding process inherently limits parallelization and makes each prediction depend only on previously generated tokens. This limitation has led to growing interest in alternative frameworks that can model and refine tokens more globally during inference. 

Diffusion LLMs~\cite{li2022diffusion, nie2025large,ye2025dream} have recently emerged as a promising alternative to autoregressive LLMs. Instead of generating text sequentially from left to right, they begin with a masked sequence and generate text with an iterative denoising process, progressively unmasking tokens over multiple steps. This enables parallel generation and allows the model to leverage bidirectional context during denoising. Recent studies~\cite{wu2025fast,li2025beyond,ma2025dkv} have increasingly focused on improving the performance of diffusion LLMs.  One promising direction is to incorporate guidance during denoising. {Motivated by the success of guidance in visual diffusion models~\cite{ho2022classifier}, recent works have incorporated guidance strategies for diffusion LLMs.} {These approaches typically introduce a weak model and adjust the prediction of the base model, using the difference between the outputs of the base and weak models.} They mainly differ in how the weak model is constructed. For example, U-CFG~\cite{nie2024scaling} replaces the prompt with masked tokens to construct a weak model, while A-CFG~\cite{li2025adaptive} forms a weak model by remasking unconfident tokens in the output of the base model. Although these approaches show that guidance can improve diffusion LLMs, they have two key limitations. {First, the influence of the weak model on the denoising process remains less clear.} In diffusion LLMs, masked tokens are prediction targets rather than null conditions as in visual diffusion~\cite{dhariwal2021diffusion,ho2022classifier}, leaving the role of the weak model unclear. As a result, guidance may {steer denoising in undesirable directions}, potentially degrading the performance. Second, these methods incur substantial inference overhead. {They use predictions of both the base and weak models, requiring two forward passes at each denoising step.} This doubles the inference cost and weakens the faster inference enabled by parallel denoising in diffusion LLMs.

{To address these limitations, we propose Global Context Guidance (GCG), a feature-space guidance method for diffusion LLMs that enhances the utilization of contextual information during denoising. {In diffusion LLMs, contextual information plays an important role in parallel denoising, as bidirectional attention allows the model to gather contextual cues from different positions across the sequence~\cite{ye2025dream,ma2026diffusion}.} To better understand this behavior, we first investigate how diffusion LLMs use contextual information during denoising. Specifically, we construct a synthetic infilling dataset that controls the distance between masked tokens and their supporting evidence in bidirectional documents, ranging from nearby evidence (\ie,~local context) to distant evidence (\ie,~global context).} {Through empirical analysis, we find that diffusion LLMs rely more heavily on local context while often underutilizing global context, suggesting that diffusion LLMs do not fully exploit the entire context available during denoising.}
{We further observe that a subset of attention heads is particularly important for leveraging global context.} {Based on this observation, GCG identifies such heads, which we refer to as global heads, and strengthens their outputs during inference.} {Specifically, GCG projects outputs of global heads onto a subspace formed by local heads~(\ie,~local space) {and amplifies the residual between the global-head outputs and their projections}, thereby increasing the contribution of global context.} Since this operation is performed directly in the feature space within a single forward pass, GCG remains efficient without requiring an additional weak model. We evaluate GCG on various models and tasks, and show that it consistently outperforms existing guidance methods, while incurring much lower inference overhead. We summarize our contributions as follows:
\begin{itemize}
\item {We present an analysis of how diffusion LLMs exploit contextual information by constructing a synthetic infilling dataset. Our analysis reveals that diffusion LLMs are biased toward local context during denoising, and a subset of attention heads plays a critical role in leveraging global context.}

\item {Based on the analysis, we propose GCG, an effective and efficient feature-space guidance method. GCG projects the outputs of selected global heads onto a local space, {and amplifies the difference between the outputs of global heads and their projections}, strengthening global-context features without requiring an additional weak model.}

\item Extensive experiments on various models and tasks show that GCG is both more effective and efficient than existing guidance methods.

\end{itemize}

\section{Related Work}
In this section, we review representative works on diffusion LLMs and guidance methods.

\subsection{Diffusion LLMs}

Diffusion LLMs~\cite{li2022diffusion, gong2022diffuseq, nie2024scaling} have emerged as an alternative paradigm to autoregressive LLMs~\cite{grattafiori2024llama, yang2025qwen3, liu2024deepseek}, generating text through iterative denoising rather than left-to-right next-token prediction. This paradigm allows parallel generation and greater flexibility in generation order, offering a promising framework for language generation. Recent large-scale models such as LLaDA~\cite{nie2025large} and Dream~\cite{ye2025dream} demonstrate that diffusion LLMs can achieve performance competitive with strong autoregressive baselines across a wide range of tasks, including general instruction following, mathematical reasoning, and code generation~\cite{clark2018arc,chen2021humaneval,cobbe2021gsm8k}.

Recent studies have explored training-free methods for improving the efficiency and effectiveness of diffusion LLMs. One line of work focuses on improving inference efficiency through KV caching. Since diffusion LLMs repeatedly compute bidirectional attention over the full sequence during denoising, standard KV caching used in autoregressive LLMs is not directly applicable. {Several works~\cite{ma2025dkv,wu2025fast} address this limitation based on the observation that the key-values of prompt and unmasked tokens change only slightly between adjacent denoising steps. They thus propose to cache the states of these tokens and reuse them across steps, reducing the computational cost.} Another line of work adapts the denoising process to use the inference budget more effectively. For example, Fast-dLLM~\cite{wu2025fast} dynamically controls the number of tokens updated at each step, while DAEDAL~\cite{li2025beyond} adjusts the output length during denoising. Recent works~\cite{wang2025remasking,hong2025wide} have also revisited the irreversible nature of standard diffusion LLMs, where tokens typically remain fixed once they are unmasked. To alleviate this limitation, these methods remask uncertain tokens and refine them again in later steps, which mitigates error propagation from incorrect early predictions.

\subsection{Guidance}

Guidance has been widely studied in diffusion models, particularly for image generation. A representative approach is classifier guidance~\cite{dhariwal2021diffusion}, which steers the denoising process using gradients from an external classifier so that generated samples better satisfy a target condition. Although effective, this approach requires a separately trained classifier and increases inference complexity. To address this limitation, classifier-free guidance (CFG)~\cite{ho2022classifier} has been proposed, which combines conditional and unconditional predictions from the diffusion model itself. CFG improves condition alignment and sample quality without requiring an additional classifier, and has become one of the most widely used guidance methods in image diffusion models. {Autoguidance~\cite{karras2024guiding} further extends this framework, introducing a weak model instead of the unconditional model, which is constructed by training with fewer iterations than the base model. It then uses the difference between the outputs of base and weak models as a guidance signal, and finds that this strengthens components that are suppressed in the weak model.} {Subsequent works have explored various ways to construct the weak model, such as perturbing the attention maps~\cite{ahn2024self} or stochastically dropping network blocks~\cite{chen2025s}, suggesting that the design of the weak model is crucial for effective guidance.}

Guidance has recently been extended from image diffusion models to diffusion LLMs. In image generation, CFG constructs an unconditional model by replacing the condition with a null prompt, and guides generation using the difference between conditional and unconditional outputs. {U-CFG~\cite{nie2024scaling} adapts this idea to diffusion LLMs, constructing a weak model, which replaces the prompt tokens with mask tokens.} A-CFG~\cite{li2025adaptive} further replaces this static weak model with a dynamic weak model, constructed by re-masking unconfident tokens in the output of the base model at each denoising step. To our knowledge, these are the only guidance methods for diffusion LLMs, showing that guidance can also improve diffusion LLMs. They, however, require an additional forward pass to obtain predictions from the weak model at each denoising step. GCG instead performs guidance directly in the feature space without constructing a weak model, remaining efficient within a single forward pass.

\section{Method}

In this section, we review the formulation of conventional diffusion LLMs~(Sec.~\ref{sec:preliminary}), analyze how these models use contextual information during denoising~(Sec.~\ref{sec:analysis}), and present our method, GCG~(Sec.~\ref{sec:GCG}).

\subsection{Preliminary}
\label{sec:preliminary}

Given a prompt sequence $\mathbf{c}$, diffusion LLMs generate a target sequence $\mathbf{x}_0$ through an iterative denoising process. Starting from a fully masked sequence, the model progressively unmasks tokens over multiple steps, where we denote by $\mathbf{x}_{\tau}$ a partially masked sequence at step $\tau$. At each step, the model takes $\mathbf{x}_{\tau}$ and $\mathbf{c}$ as input, and predicts the original tokens at the masked positions as follows:
\begin{equation}
p_{\theta}(\mathbf{x}_0 \mid \mathbf{x}_{\tau}, \mathbf{c}).
\end{equation}
Existing guidance methods for diffusion LLMs typically introduce a weak model in addition to the base model. Let $\mathbf{z}_{\mathrm{base}}(i,\tau)$ and $\mathbf{z}_{\mathrm{weak}}(i,\tau)$ denote the logits at token position $i$ and denoising step $\tau$ from the base and weak models, respectively. Guidance is then applied at the logit level as
\begin{equation}
\mathbf{z}_{\mathrm{guided}}(i,\tau)
=
\mathbf{z}_{\mathrm{base}}(i,\tau)
+
s \left( \mathbf{z}_{\mathrm{base}}(i,\tau) - \mathbf{z}_{\mathrm{weak}}(i,\tau) \right),
\end{equation}
where $s$ is a guidance scale. This formulation increases the contribution of directions that are suppressed in the weak model, but it requires an additional forward pass to obtain $\mathbf{z}_{\mathrm{weak}}(i,\tau)$.

\begin{figure}[t]
\centering
\includegraphics[width=0.75\textwidth]{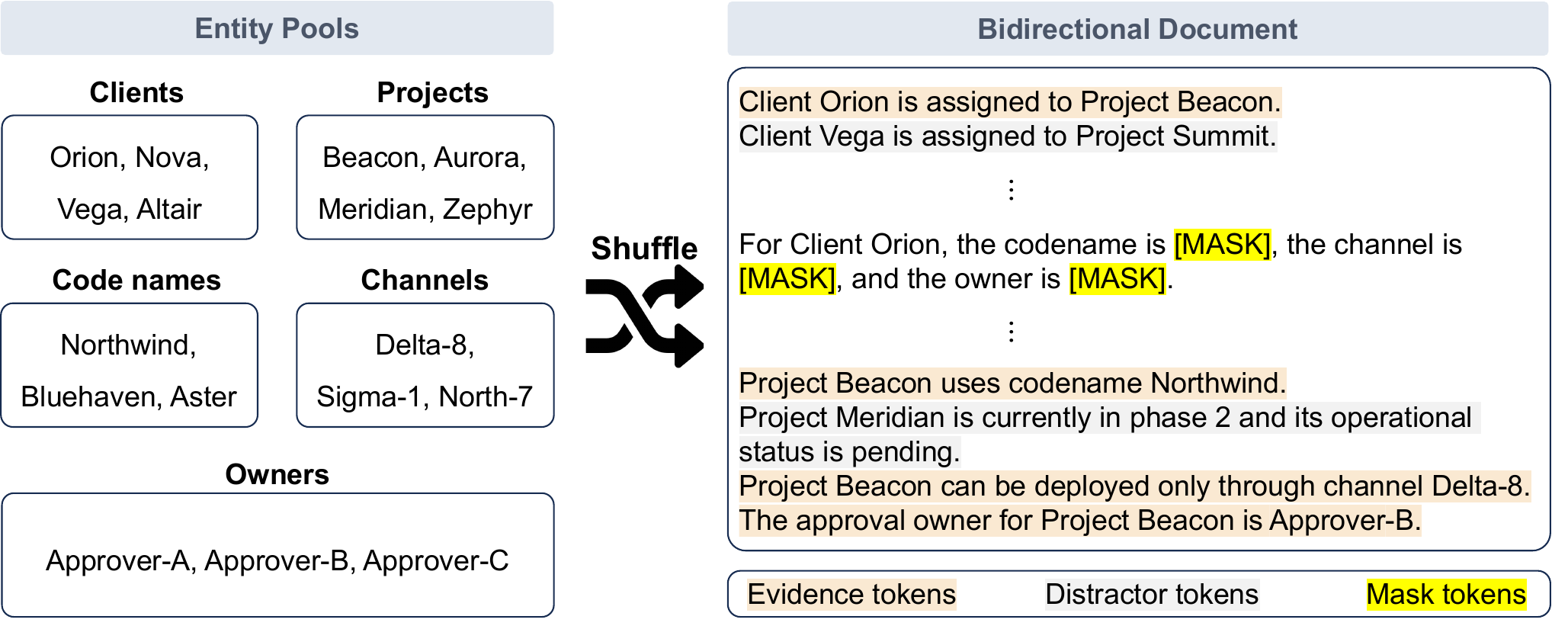}
\caption{Data generation process for a synthetic infilling benchmark. Entities are randomly sampled from predefined pools and used to construct a bidirectional document. In the resulting document, supporting evidence is placed on both sides of masked tokens, and separated from them by distractor tokens. {The distances from the masked tokens to the supporting evidence are varied randomly for each sample.}}
\label{fig:benchmark}
\end{figure}

\subsection{Analysis}
\label{sec:analysis}

\subsubsection{Contextual-distance analysis}
To analyze how diffusion LLMs leverage contextual information during denoising, we construct a synthetic infilling benchmark~(Fig.~\ref{fig:benchmark}). Specifically, we generate each sample through randomly sampling entities from predefined pools and using them to construct a bidirectional document. The resulting document contains masked tokens in the middle and supporting evidence on both sides. By randomly varying the distances from the masked tokens to the left and right evidence, we obtain examples spanning a broad range of local and global-context settings. {For more details on the data generation process, please refer to Appendix~\ref{sec:app_synthetic}.}

Using this benchmark, we analyze the effect of evidence--mask distance on the performance of LLaDA-8B~\cite{nie2025large}. We can see from the gray line of Fig.~\ref{fig:baseline_distance} that the performance degrades consistently as the evidence becomes more distant from the masked tokens. This suggests that, although diffusion LLMs can access the entire context during denoising, they are biased toward local context and underutilize global context.

\subsubsection{Head-level analysis}

We next analyze attention patterns of individual heads using the center of mass (COM) of their attention distributions. Let $\alpha_h^m(i,j,\tau)$ denote the attention weight from query token $i$ to key token $j$ at block $m$, head $h$, and denoising step $\tau$. For each head, we compute the average distance between query and key tokens, where the distance $|i-j|$ is weighted by the corresponding attention weight~$\alpha_h^m(i,j,\tau)$:
\begin{equation}
\frac{1}{T}
\sum_{\tau=1}^{T}
\left(
\frac{1}{N}
\sum_i \sum_j \alpha_h^m(i,j,\tau) |i-j|
\right), 
\end{equation}
where $T$ is the number of denoising steps and $N$ is the number of query tokens. We refer to this value as the COM of head $h$ in block $m$. Larger COM scores indicate that the head attends more strongly to distant positions on average, and vice versa.

\begin{figure}[t]
  \small
\centering
\begin{subfigure}[t]{0.38\textwidth}
  \centering
  \includegraphics[height=3.5cm]{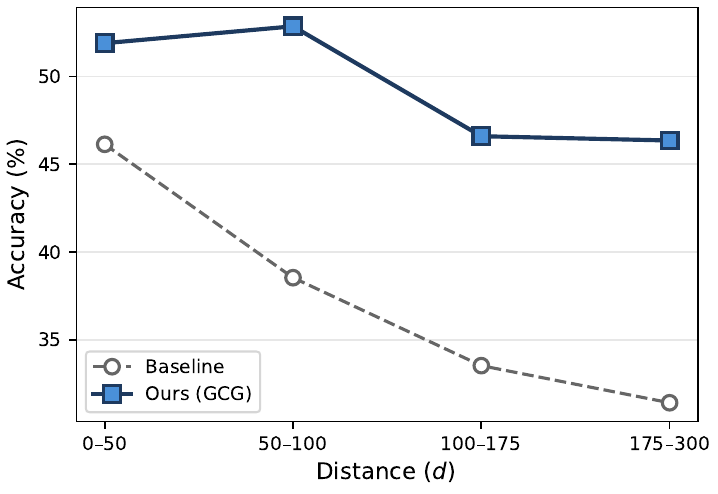}
  \caption{Accuracy vs. Evidence-Mask Distance.}
  \label{fig:baseline_distance}
\end{subfigure}
\hfill
\begin{subfigure}[t]{0.58\textwidth}
  \centering
  \includegraphics[height=3.5cm]{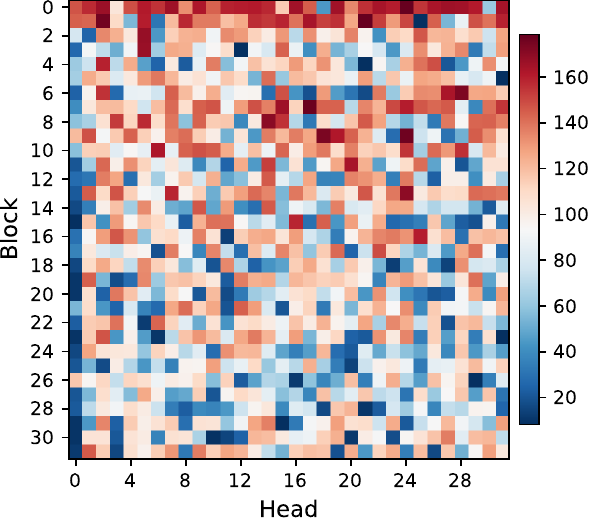}
  \caption{Average COM of attention heads across blocks.}
  \label{fig:com_heatmap}
\end{subfigure}
\caption{(a) Accuracy as a function of the average evidence--mask distance. The baseline performance consistently degrades as the distance increases, indicating that diffusion LLMs rely more heavily on local context than on global context during denoising (gray). GCG mitigates this issue by enhancing global-context utilization, yielding larger gains at longer distances (blue). (b) COM heatmap across all blocks and heads. Heads with high COM values (red) attend more strongly to global context, whereas heads with low COM values (blue) mainly focus on local context.}
\label{fig:analysis}
\end{figure}

As shown in Fig.~\ref{fig:com_heatmap}, attention heads exhibit substantial variation in their average COM values. Some heads have relatively small COM values and primarily attend to nearby positions, whereas others have much larger values and attend more strongly to distant positions. To examine whether this tendency is preserved across samples, we compute head rankings induced by COM for each sample and measure their pairwise Spearman correlation. The resulting rankings are highly consistent, with a mean correlation of {0.94 across 700 samples}. 

\begin{figure}[t]
  \small
\centering
\begin{subfigure}[t]{0.3\textwidth}
  \centering
  \includegraphics[height=2.5cm,width=\linewidth,keepaspectratio]{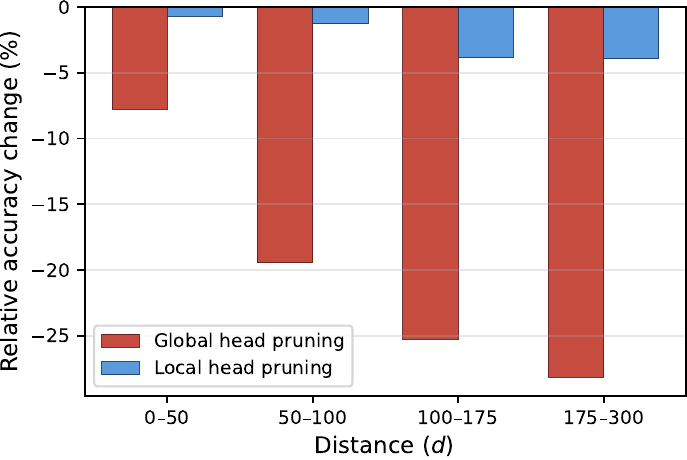}
  \caption{Global vs.\ local head pruning.}
  \label{fig:prune_com}
\end{subfigure}
\hfill
\begin{subfigure}[t]{0.55\textwidth}
  \centering
  \includegraphics[height=2.5cm,width=\linewidth,keepaspectratio]{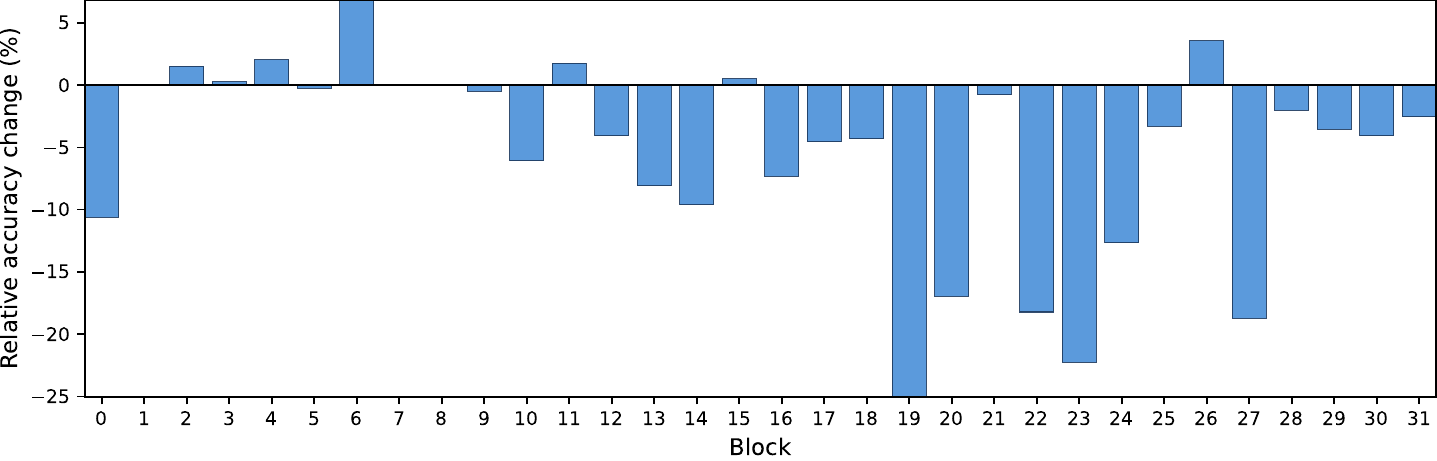}
  \caption{Block-wise pruning of global heads.}
  \label{fig:layer_knockout}
\end{subfigure}
\hfill
\caption{(a) Relative accuracy change after pruning global heads (red) or local heads (blue) across evidence--mask distances. Pruning global heads leads to larger degradation, especially at longer distances. (b) Block-wise relative accuracy change after removing global heads from each block individually. The impact varies substantially across blocks, indicating that a subset of blocks is particularly  more critical for global-context utilization.}
\label{fig:head_importance}
\end{figure}

{We then examine whether heads with high COM values are important for leveraging distant context. For this analysis, we rank heads in each block by their COM values, and refer to the top-25\% and bottom-25\% heads as global and local heads, respectively.} We prune either global or local heads by zeroing the outputs of the corresponding heads during inference, and measure the resulting performance change on the synthetic infilling benchmark. As shown in Fig.~\ref{fig:prune_com}, pruning global heads causes substantially larger performance degradation than pruning local heads, and this gap becomes more pronounced as the evidence--mask distance increases. This indicates that global heads play a more important role in leveraging distant context during denoising.

To further analyze whether all global heads are equally important across blocks, we perform a block-wise ablation study, where global heads are removed from one block at a time, while keeping those in all other blocks unchanged. As shown in Fig.~\ref{fig:layer_knockout}, the impact varies substantially across blocks, indicating that the contribution of global heads is not uniform across blocks. Additional analyses on Dream-7B~\cite{ye2025dream} and LLaDA-1.5~\cite{zhu2025llada} are provided in Appendix~\ref{sec:app_results}, showing similar trends.

\subsection{Global Context Guidance (GCG)}
\label{sec:GCG}

Based on the above observations, we propose GCG, a feature-space guidance method for diffusion LLMs that enhances global-context utilization during denoising. GCG identifies target blocks for global-context utilization, using COM analysis and block-wise ablation, and applies guidance to global heads in these blocks. {For each selected global head, GCG projects its output onto the subspace spanned by outputs of local heads in the same block. It then amplifies the residual between the outputs of global heads and their projection, thereby strengthening global-context features during denoising.}

{We identify global and local heads and target blocks for GCG using a training subset of LongBench~\cite{bai2024longbench}\footnote{We use a LongBench subset rather than the synthetic benchmark for head and block selection to reflect more realistic generation settings. To build the subset, we randomly sample 3 examples from each of the 21 tasks, including question answering, summarization, code, and dialogue, resulting in a total of 63 samples.}. For each model, we first compute COM values on this subset and rank heads within each block, and designate top-$p$\% heads as global heads and the bottom-$p$\% heads as local heads. We then perform block-wise ablation, pruning global heads in each block and select the top-$q$\% blocks with the largest performance drops as target blocks. The selected heads and target blocks are fixed for all evaluation benchmarks.}

Let $\mathcal{M}$ denote the set of target blocks. For each target block $m \in \mathcal{M}$, let
$\mathcal{G}^m = \{g_1,\ldots,g_k\}$ and $\mathcal{L}^m = \{j_1,\ldots,j_k\}$
denote the sets of selected global and local heads, respectively, where $k$ is the number of heads in each set.
We denote by $\mathbf{o}_h^m(i,\tau) \in \mathbb{R}^{D_h}$ the output of head $h$ at token position $i$ and denoising step $\tau$ in block $m$, where $D_h$ is the head dimension. We construct the local space from the outputs of local heads:
\begin{equation}
\mathbf{A}^m(i,\tau)
=
\left[
\mathbf{o}_{j_1}^m(i,\tau)^\top;\,
\mathbf{o}_{j_2}^m(i,\tau)^\top;\,
\ldots;\,
\mathbf{o}_{j_k}^m(i,\tau)^\top
\right]
\in \mathbb{R}^{k \times D_h}.
\end{equation}
The row space of $\mathbf{A}^m(i,\tau)$ defines a local space at token position $i$ and denoising step $\tau$. For each selected global head {$g_l \in \mathcal{G}^m$}, {we compute a projection of its output onto the local space~\cite{golub2013matrix}:}
\begin{equation}
\mathrm{Proj}_{\mathcal{L}^m}\!\left(\mathbf{o}_{g_l}^m(i,\tau)\right)
=
\mathbf{A}^m(i,\tau)^\top
\left(
\mathbf{A}^m(i,\tau)\mathbf{A}^m(i,\tau)^\top + \lambda \mathbf{I}
\right)^{-1}
\mathbf{A}^m(i,\tau)\mathbf{o}_{g_l}^m(i,\tau),
\label{eq:proj}
\end{equation}
where $\lambda$ is a small constant for numerical stability, which we set to $10^{-4}$ in our experiments. This projection is efficient, since it can be computed by solving the corresponding $k \times k$ linear system, instead of inverting the matrix in Eq.~\eqref{eq:proj}~\cite{golub2013matrix}. Since $k$ is the number of local heads in the block, which is typically small (\eg, 8 in LLaDA-8B~\cite{nie2025large}), this adds negligible overhead during inference.
Note that the residual
\begin{equation}
\mathbf{o}_{g_l}^m(i,\tau) - \mathrm{Proj}_{\mathcal{L}^m}\!\left(\mathbf{o}_{g_l}^m(i,\tau)\right)
\end{equation}
removes the component of the global-head output aligned with local-head features, {capturing the features that are more related to global context}. GCG amplifies this residual, and updates the output of each selected global head:
\begin{equation}
\tilde{\mathbf{o}}_{g_l}^m(i,\tau)
=
\mathbf{o}_{g_l}^m(i,\tau)
+
s
\left(
\mathbf{o}_{g_l}^m(i,\tau)
-
\mathrm{Proj}_{\mathcal{L}^m}\!\left(\mathbf{o}_{g_l}^m(i,\tau)\right)
\right),
\label{eq:guidance}
\end{equation}
where $s$ is a guidance scale shared across all target heads and blocks. Outputs of all other heads are left unchanged. This update can be viewed as an implicit guidance mechanism without an explicit weak model. Rather than contrasting outputs from separate models, GCG directly modifies head-wise features within the same model, requiring only a single forward pass, while enhancing the performance significantly~(blue line of Fig.~\ref{fig:baseline_distance}). GCG is also compatible with FlashAttention~\cite{dao2022flashattention}, since  it operates on the features after attention computation.

\section{Experiments}
\label{sec:experiments}
In this section, we first describe the experimental settings (Sec.~\ref{sec:experimental_settings}). We then provide a quantitative comparison of GCG with existing guidance methods across various models and benchmarks (Sec.~\ref{sec:quantitative}), and present an empirical analysis of GCG (Sec.~\ref{sec:exp_analysis}).

\subsection{Experimental settings}
\label{sec:experimental_settings}

\textbf{Models and benchmarks.}
We evaluate GCG on three representative diffusion LLMs, including LLaDA-8B~\cite{nie2025large}, Dream-7B~\cite{ye2025dream}, and LLaDA-1.5~\cite{zhu2025llada}. We use a diverse set of benchmarks covering scientific reasoning (ARC~\cite{clark2018arc}), physical commonsense reasoning (PIQA~\cite{bisk2020piqa}), mathematical reasoning (GSM8K-CoT~\cite{cobbe2021gsm8k}), code generation (HumanEval~\cite{chen2021humaneval} and MBPP~\cite{austin2021mbpp}), and long-context understanding (Ruler~\cite{hsieh2024ruler} and MiniLongBench~\cite{huang2025minilongbench}). Following the standard evaluation protocols of each benchmark, we report accuracy for ARC,  PIQA, and GSM8K-CoT, and pass@1 for HumanEval and MBPP. For Ruler and MiniLongBench, we report the average score over all subtasks. We compare GCG with representative guidance methods for diffusion LLMs, including U-CFG~\cite{nie2024scaling} and A-CFG~\cite{li2025adaptive}, which we reproduce based on the descriptions in their papers. 

\textbf{Implementation details.}
As described in Sec.~\ref{sec:GCG}, we use a training subset of LongBench~\cite{bai2024longbench} to select global and local heads and target blocks. Specifically, heads in each block are ranked by COM scores, where the top-$p$\% and bottom-$p$\% heads are selected as global and local heads, respectively. Target blocks are selected as the top-$q$\% blocks with the largest performance drops after performing block-wise pruning of global heads, and these heads and blocks are fixed for each model. Unless otherwise specified, we set both $p$ and $q$ to 25, and the guidance scale $s$ in Eq.~(\ref{eq:guidance}) to 0.3. All experiments are performed on 8 A6000 GPUs.
Further implementation details, including decoding settings, are provided in Appendix~\ref{sec:app_imp}.

\begin{table}[t]
  \small
\centering
\caption{Quantitative comparison with existing guidance methods across models and benchmarks. We evaluate LLaDA-8B~\cite{nie2025large}, Dream-7B~\cite{ye2025dream}, and LLaDA-1.5~\cite{zhu2025llada} on ARC~\cite{clark2018arc}, PIQA~\cite{bisk2020piqa}, HumanEval~\cite{chen2021humaneval}, MBPP~\cite{austin2021mbpp}, GSM8K-CoT~\cite{cobbe2021gsm8k}, Ruler~\cite{hsieh2024ruler}, and MiniLongBench~\cite{huang2025minilongbench}. We also report the inference overhead of each method in terms of FLOPs, measured for a single denoising step with a context length of 1K tokens. Bold numbers indicate the best performance within each model.}
\label{tab:benchmark}
\setlength{\tabcolsep}{3pt}
\resizebox{\textwidth}{!}{%
\begin{tabular}{lcccccccc}
\toprule
Method & ARC & PIQA & HumanEval & MBPP & GSM8K-CoT & Ruler & MiniLongBench & FLOPs \\
\midrule
LLaDA-8B~\cite{nie2025large} & 43.3 & 72.4 & 12.2 & 16.0 & 46.7 & 26.2 & 24.8 & \textbf{15.9T} \\
+ U-CFG~\cite{nie2024scaling} & 42.4 & 74.1 & ~~7.9 & 14.8 & 36.7 & 28.9 & 22.8 & 31.8T \\
+ A-CFG~\cite{li2025adaptive} & 45.7 & 73.8 & 11.6 & 16.6 & 34.7 & 27.5 & 24.4 & 31.8T \\
\rowcolor{cyan!8}
+ GCG & \textbf{46.2} & \textbf{74.4} & \textbf{12.8} & \textbf{18.4} & \textbf{48.4} & \textbf{29.5} & \textbf{25.9} & \textbf{15.9T} \\
\midrule
Dream-7B~\cite{ye2025dream} & 54.2 & 75.5 & ~~1.8 & 15.0 & ~~3.9 & 28.0 & 26.5 & \textbf{14.9T} \\
+ U-CFG~\cite{nie2024scaling} & 55.6 & 74.8 & ~~3.1 & 14.6 & ~~4.3 & 28.3 & 26.7 & 29.8T \\
+ A-CFG~\cite{li2025adaptive} & 56.1 & 74.7 & ~~\textbf{3.7} & 12.6 & ~~5.2 & 27.1 & 26.8 & 29.8T \\
\rowcolor{cyan!8}
+ GCG & \textbf{58.4} & \textbf{75.6} & ~~3.1 & \textbf{17.0} & ~~\textbf{5.4} & \textbf{28.5} & \textbf{27.1} & \textbf{14.9T} \\
\midrule
LLaDA-1.5~\cite{zhu2025llada} & 51.4 & 73.5 & 15.9 & 17.2 & 37.8 & 37.6 & 31.6 & \textbf{15.9T} \\
+ U-CFG~\cite{nie2024scaling} & 53.8 & 73.2 & 18.1 & \textbf{19.6} & 38.2 & 38.2 & 29.2 & 31.8T \\
+ A-CFG~\cite{li2025adaptive} & \textbf{54.5} & 73.8 & 17.4 & 18.4 & 37.9 & 38.4 & 29.4 & 31.8T \\
\rowcolor{cyan!8}
+ GCG & \textbf{54.5} & \textbf{73.9} & \textbf{18.9} & 19.0 & \textbf{38.5} & \textbf{39.3} & \textbf{32.2} & \textbf{15.9T} \\
\bottomrule
\end{tabular}%
}
\end{table}

\subsection{Results}
\label{sec:quantitative}

We show in Table~\ref{tab:benchmark} quantitative comparisons between GCG and existing guidance methods. We summarize our findings as follows: (1) GCG consistently improves baselines across various models and benchmarks. For example, on LLaDA-8B~\cite{nie2025large}, GCG improves GSM8K-CoT~\cite{cobbe2021gsm8k} from 46.7 to 48.4 and Ruler from 26.2 to 29.5. This suggests that enhancing global-context utilization during denoising is broadly effective, rather than being specific to a particular model or task. (2) GCG achieves stronger overall performance than existing guidance methods. While U-CFG~\cite{nie2024scaling} and A-CFG~\cite{li2025adaptive} improve the baseline in some cases, their gains are less consistent, and they can even degrade performance on several tasks, such as GSM8K-CoT on LLaDA-8B. Note that these methods construct a weak model by replacing prompt or generated tokens with masked tokens, analogous to using a null prompt in image diffusion models~\cite{ho2022classifier}. However, masked tokens in diffusion LLMs are prediction targets rather than null conditions, making the role of the weak model less clear. GCG instead builds on our analysis of context utilization in diffusion LLMs and applies guidance to components that are important for leveraging global context, leading to more consistent improvements across tasks. (3) GCG is more efficient than existing guidance methods. Since U-CFG and A-CFG require an additional forward pass for the weak model, they substantially increase the inference cost. GCG avoids this overhead by modifying head features within a single forward pass, which requires only negligible overhead (\eg,~0.003\% FLOPs increase for LLaDA-8B). These results demonstrate that GCG is both effective and efficient compared to existing guidance methods for diffusion LLMs.

\begin{table}[t]
  \small
\centering
\caption{Quantitative comparison on the design choices of GCG across models and benchmarks. LCG replaces global-head guidance with local-head guidance, while LSB applies GCG to the least sensitive blocks instead of the selected target blocks. Bold numbers indicate the best performance within each model. Please refer to the main text for details.}
\label{tab:analysis_ablation}
\setlength{\tabcolsep}{3pt}
\resizebox{0.85\textwidth}{!}{%
\begin{tabular}{lccccccc}
\toprule
Method & ARC & PIQA & HumanEval & MBPP & GSM8K-CoT & Ruler & MiniLongBench \\
\midrule
LLaDA-8B~\cite{nie2025large} 
& 43.3 & 72.4 & 12.2 & 16.0 & 46.7 & 26.2 & 24.8 \\
+ LCG 
& 45.2 & 73.7 & 11.0 & 16.8 & 47.6 & 26.0 & 25.0 \\
+ LSB
& 45.3 & 73.9 & 11.0 & 16.6 & 44.8 & 23.4 & 24.6 \\
\rowcolor{cyan!8}
+ GCG 
& \textbf{46.2} & \textbf{74.4} & \textbf{12.8} & \textbf{18.4} & \textbf{48.4} & \textbf{29.5} & \textbf{25.9} \\
\midrule
Dream-7B~\cite{ye2025dream} 
& 54.2 & 75.5 & 1.8 & 15.0 & 3.9 & 28.0 & 26.5 \\
+ LCG 
& 57.3 & 75.4 & 1.8 & 13.8 & 2.7 & 22.4 & 25.6 \\
+ LSB
& 58.0 & 75.5 & 2.4 & 14.4 & 3.3 & 25.9 & 26.1 \\
\rowcolor{cyan!8}
+ GCG 
& \textbf{58.4} & \textbf{75.6} & \textbf{3.1} & \textbf{17.0} & \textbf{5.4} & \textbf{28.5} & \textbf{27.1} \\
\midrule
LLaDA-1.5~\cite{zhu2025llada} 
& 51.4 & 73.5 & 15.9 & 17.2 & 37.8 & 37.6 & 31.6 \\
+ LCG 
& 51.0 & 73.6 & 15.2 & 16.0 & 29.8 & 37.2 & 31.4 \\
+ LSB
& 52.0 & 73.6 & 16.5 & 17.2 & 31.2 & 37.8 & 31.4 \\
\rowcolor{cyan!8}
+ GCG 
& \textbf{54.5} & \textbf{73.9} & \textbf{18.9} & \textbf{19.0} & \textbf{38.5} & \textbf{39.3} & \textbf{32.2} \\
\bottomrule
\end{tabular}%
}
\vspace{-0.2cm}
\end{table}

\subsection{Analysis}
\label{sec:exp_analysis}

\begin{table}[t]
\centering
\caption{Quantitative analysis of GCG with different ratios for selecting global and local heads ($p$) and target blocks ($q$) on LLaDA-8B~\cite{nie2025large}. Note that it is equivalent to the baseline without guidance when both $p$ and $q$ are set to 0. Best scores in each column are in bold.}
\label{tab:hyperparam_llada8b}
\resizebox{0.9\linewidth}{!}{
\begin{tabular}{ccccccccc}
\toprule
$p$ & $q$ & ARC & PIQA & HumanEval & MBPP & GSM8K-CoT & Ruler & MiniLongBench \\
\midrule
0.0  & 0.0  & 43.3 & 72.4 & 12.2 & 16.0 & 46.7 & 26.2 & 24.8 \\
\midrule
\multirow{3}{*}{12.5}
& 12.5 & 45.7 & 73.9 & \textbf{13.4} & 17.8 & \textbf{49.5} & 28.1 & 25.6 \\
& 25.0 & 44.9 & 73.9 & 12.2 & \textbf{18.8} & 49.1 & 28.3 & 25.0 \\
& 50.0 & 44.7 & 74.3 & 11.6 & 17.6 & 49.3 & 26.0 & 26.6 \\
\cmidrule(lr){1-9}
\multirow{3}{*}{25.0}
& 12.5 & 45.7 & 73.7 & \textbf{13.4} & 18.2 & 46.8 & 30.2 & 25.9 \\
& 25.0 & \textbf{46.2} & \textbf{74.4} & 12.8 & 18.4 & 48.4 & 29.5 & 25.9 \\
& 50.0 & 45.0 & 74.1 & 12.8 & 16.6 & 47.3 & 29.1 & 27.0 \\
\cmidrule(lr){1-9}
\multirow{3}{*}{50.0}
& 12.5 & 45.5 & 73.9 & 12.2 & 17.4 & 47.5 & 30.8 & 25.2 \\
& 25.0 & 45.1 & 74.0 & 12.2 & 18.0 & 47.3 & \textbf{30.9} & 24.8 \\
& 50.0 & 44.1 & 73.9 & 12.8 & 17.0 & 47.7 & 29.2 & \textbf{27.7} \\
\bottomrule
\end{tabular}
}
\end{table}

\begin{table}[t]
\centering
\caption{Quantitative analysis of GCG with different guidance scale $s$ on LLaDA-8B~\cite{nie2025large}. Note that it is equivalent to the baseline without guidance when $s$ is set to 0. Best scores in each column are in bold.}
\label{tab:hyperparam_s_llada8b}
\resizebox{0.8\linewidth}{!}{
\begin{tabular}{cccccccc}
\toprule
$s$ & ARC & PIQA & HumanEval & MBPP & GSM8K-CoT & Ruler & MiniLongBench \\
\midrule
0.0 & 43.3 & 72.4 & 12.2 & 16.0 & 46.7 & 26.2 & 24.8 \\
0.1 
& 46.0 & 74.1 & \textbf{12.8} & 18.0 & 47.0 & 28.5 & 24.9 \\
0.2 
& 45.7 & 74.3 & \textbf{12.8} & 18.8 & \textbf{50.6} & 28.6 & 25.6 \\
0.3 
& \textbf{46.2} & 74.4 & \textbf{12.8} & 18.4 & 48.4 & \textbf{29.5} & \textbf{25.9} \\
0.4 
& \textbf{46.2} & 74.4 & 12.2 & 18.0 & 48.1 & 27.6 & 25.0 \\
0.5 
& 46.0 & \textbf{74.5} & 12.2 & \textbf{19.4} & 48.1 & 25.7 & 24.5 \\
\bottomrule
\end{tabular}
}
\end{table}

\textbf{Design choices of GCG.}
We analyze in Table~\ref{tab:analysis_ablation} two design choices of GCG. We first analyze whether guidance should enhance global-context features rather than local-context features. To examine this, we construct local context guidance (LCG), a variant that reverses the direction of GCG. LCG projects outputs of local heads onto the global-head subspace and amplifies the difference between the original and projected outputs. We can see that LCG shows worse performance than GCG. This is particularly evident on benchmarks that require chain-of-thought reasoning~(\eg, GSM8K-CoT), or long-context understanding~(\eg, Ruler and MiniLongBench), which require more global-context utilization. This is consistent with our analysis in Sec.~\ref{sec:analysis}, where diffusion LLMs already rely heavily on local context during denoising. Further strengthening local-context features can therefore reinforce this bias, reducing the benefit of guidance. We further analyze the effect of target block selection. Instead of selecting the 25\% blocks with the largest performance drops after block-wise pruning of global heads as in GCG, we apply the same global-head guidance operation to the 25\% blocks with the smallest performance drops, which we refer to this variant as Least-Sensitive Blocks (LSB). Similar to LCG, LSB also shows worse performance than GCG, especially on benchmarks that require more global-context utilization. This suggests that global-head guidance is effective when applied to blocks that are important for global-context utilization.

\textbf{Hyperparameters.}
We analyze the sensitivity of GCG to the hyperparameters used for selecting heads and blocks and for controlling the guidance strength. We show in Table~\ref{tab:hyperparam_llada8b} the results obtained by sweeping $p,q \in \{12.5,25,50\}$, where $p$ and $q$ denote the percentages of selected heads and target blocks, respectively. We also show in Table~\ref{tab:hyperparam_s_llada8b} the results obtained by sweeping the guidance scale $s \in \{0.1,0.2,0.3,0.4,0.5\}$. We can see that GCG consistently improves the baseline across a wide range of settings, indicating that it is robust to the choice of hyperparameters. This suggests that the effectiveness of GCG does not rely on a carefully tuned configuration, but rather comes from enhancing global-context features during denoising. Please refer to Appendix~\ref{sec:app_results} for additional results on Dream-7B~\cite{ye2025dream} and LLaDA-1.5~\cite{zhu2025llada}.

\begin{figure}[t]
\small
\centering
\includegraphics[width=0.95\textwidth]{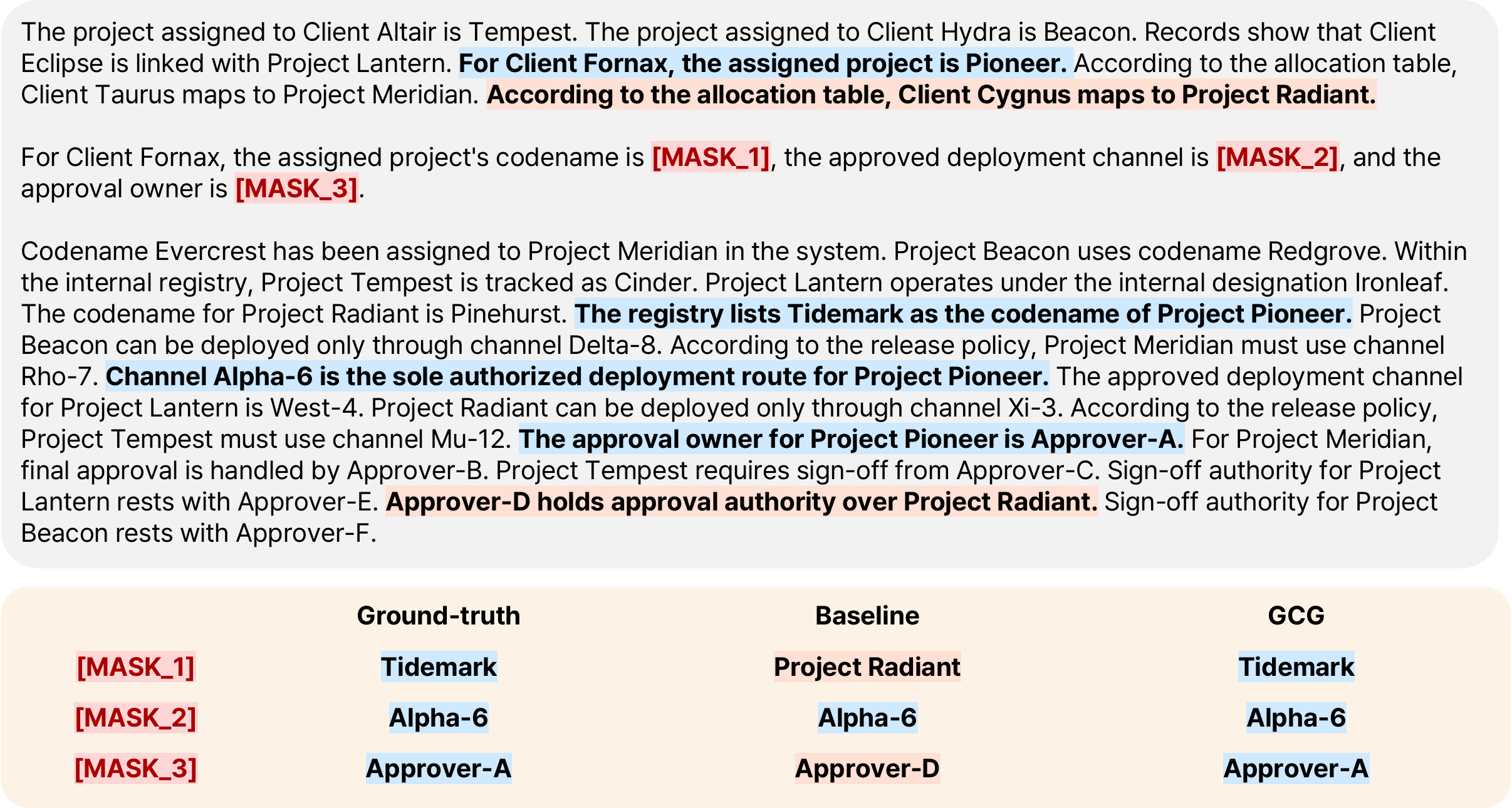}
\caption{Qualitative example on LLaDA-1.5~\cite{zhu2025llada} using the synthetic infilling benchmark. Blue highlights indicate supporting evidence for the correct predictions, orange highlights indicate distractor tokens used by the baseline.}
\label{fig:qualitative}
\end{figure}

\textbf{Qualitative analysis.}
We show in Fig.~\ref{fig:qualitative} a qualitative example on LLaDA-1.5~\cite{zhu2025llada} using the synthetic infilling dataset, illustrated in Fig.~\ref{fig:benchmark}. In this dataset, the model needs to recover the masked tokens by identifying the relevant evidence in the document. We can see that the baseline fails on recovering several masked tokens. Specifically, the baseline either relies on nearby distractor tokens~(\eg, predicting Project Radiant for \texttt{[MASK\_1]}) or incorrectly uses irrelevant distant evidence~(\eg, predicting Approver-D for \texttt{[MASK\_3]}). This suggests that the baseline can be misled by both local distractors and irrelevant global context. In contrast, GCG correctly recovers all masked tokens, indicating that it identifies and uses the relevant contextual information more effectively.

\paragraph{Calibration data.}
GCG selects global and local heads and target blocks using a subset of LongBench~\cite{bai2024longbench} (Sec.~3.3).
To analyze whether GCG depends on this choice, we perform the selection process using different calibration sets, including GSM8K-CoT~\cite{cobbe2021gsm8k}, HumanEval~\cite{chen2021humaneval}, and Ruler~\cite{hsieh2024ruler}, while keeping all other settings fixed.
We show in Table~\ref{tab:calib} the results on LLaDA-8B~\cite{nie2025large}.
We can see that GCG consistently outperforms the baseline for all calibration sets, and the results obtained with different calibration sets are comparable to each other.
This suggests that GCG is robust to the choice of calibration data.

\paragraph{Applicability to other transformers.}
GCG operates on the outputs of attention heads, and it does not depend on the training or inference procedure of diffusion LLMs.
It can thus be applied to other transformer-based language models, such as autoregressive LLMs and bidirectional masked language models.
We conjecture, however, that GCG is particularly beneficial for diffusion LLMs.
At early denoising steps, many tokens near a masked position remain masked, and the local context provides limited information for the prediction.
The prediction should thus rely more on distant evidence, making the bias toward local context more harmful in diffusion LLMs.
In contrast, autoregressive LLMs predict each token from a fully observed prefix, and masked language models typically predict masked tokens in a single step with most surrounding tokens observed.
To verify this, we apply GCG to Llama-3.2-3B~\cite{grattafiori2024llama} and RoBERTa-base~\cite{liu2019roberta}, and show in Table~\ref{tab:llama} the results with different guidance scales.
We can see that GCG improves the performance on several tasks for both models, while the gains are smaller and less consistent than those for diffusion LLMs.
This suggests that GCG is applicable to general transformers, while more effective for diffusion LLMs, where the local context is sparse during denoising.

\begin{table}[t]
  \centering
  \caption{Quantitative results of GCG on LLaDA-8B~\cite{nie2025large} with different calibration sets. Each row denotes the dataset used to select global heads, local heads, and target blocks. We report accuracy for ARC~\cite{clark2018arc}, PIQA~\cite{bisk2020piqa}, and GSM8K-CoT~\cite{cobbe2021gsm8k}, pass@1 for HumanEval~\cite{chen2021humaneval} and MBPP~\cite{austin2021mbpp}, and average scores for Ruler~\cite{hsieh2024ruler} and MiniLongBench~\cite{huang2025minilongbench}. Numbers in bold indicate the best performance.}
  \label{tab:calib}
  \small
  \setlength{\tabcolsep}{4pt}
  \begin{tabular}{lccccccc}
    \toprule
    Calibration set & ARC & PIQA & HumanEval & MBPP & GSM8K-CoT & Ruler & MiniLongBench \\
    \midrule
    LLaDA-8B~\cite{nie2025large}             & 43.3 & 72.4 & 12.2 & 16.0 & 46.7 & 26.2 & 24.8 \\
    \midrule
    GSM8K-CoT~\cite{cobbe2021gsm8k}          & 45.7 & \textbf{74.6} & 12.8 & 18.0 & \textbf{48.6} & 28.7 & 25.4 \\
    HumanEval~\cite{chen2021humaneval}       & 45.6 & 74.1 & \textbf{13.4} & 18.2 & 47.9 & 28.5 & \textbf{26.1} \\
    Ruler~\cite{hsieh2024ruler}              & 45.8 & 74.0 & 12.8 & 18.2 & 48.0 & 29.2 & 25.7 \\
    LongBench~\cite{bai2024longbench} (default) & \textbf{46.2} & 74.4 & 12.8 & \textbf{18.4} & 48.4 & \textbf{29.5} & 25.9 \\
    \bottomrule
  \end{tabular}
\end{table}

\begin{table}[t]
  \centering
  \caption{GCG on Llama-3.2-3B~\cite{grattafiori2024llama} and RoBERTa-base~\cite{liu2019roberta} across guidance scales $s$. For RoBERTa-base, we report development-set F1 on SQuAD v2~\cite{rajpurkar2018squad} and the average development-set score on GLUE~\cite{wang2019glue}. Best scores in each column are in bold.}
  \label{tab:llama}
  \label{tab:roberta}
  \small
  \renewcommand{\arraystretch}{1.1}
  \setlength{\tabcolsep}{5pt}
  \begin{tabular}{@{}c cccc c cc@{}}
    \toprule
\multirow{2}{*}{$s$}
  & \multicolumn{4}{c}{Llama-3.2-3B}
  && \multicolumn{2}{c}{RoBERTa-base} \\
\cmidrule(lr){2-5}\cmidrule(l){7-8}
  & ARC-c & HumanEval & MBPP & MiniLongBench
  && SQuAD v2 & GLUE \\
\midrule
    0.0 & 46.1 & 51.2 & 52.4 & 21.8 && 82.92 & \textbf{84.70} \\
    0.1 & 46.0 & \textbf{51.8} & 52.0 & 21.1 && 83.20 & 84.39 \\
    0.2 & \textbf{46.6} & \textbf{51.8} & \textbf{53.0} & 21.2 && 83.36 & 84.36 \\
    0.3 & 46.0 & \textbf{51.8} & 52.0 & 21.4 && \textbf{83.44} & 84.66 \\
    0.4 & 46.2 & \textbf{51.8} & 50.6 & \textbf{21.9} && 83.37 & 84.45 \\
    0.5 & 45.9 & 50.6 & 52.0 & 21.6 && 82.96 & 84.00 \\
    \bottomrule
  \end{tabular}
\end{table}

\section{Limitations and Future Work}
\label{sec:limitation}

GCG requires a separate selection process to identify global and local heads and target blocks. In our experiments, this process takes about 2--3 hours per model on 8 NVIDIA A6000 GPUs. Note that the selection is performed only once for each model, and the selected heads and blocks are fixed across all benchmarks. Nevertheless, improving this selection process would further enhance the practicality of GCG, especially for larger diffusion LLMs. Future work could reduce the required number of samples for selection or replace block-wise ablation with lightweight proxy metrics for identifying target blocks.

\section{Conclusion}

We have presented an empirical analysis of how diffusion LLMs use contextual information during denoising. Our analysis shows that diffusion LLMs tend to rely more heavily on local context, while often underutilizing global context. Based on this observation, we have introduced GCG, a feature-space guidance method that enhances global-context utilization without constructing an explicit weak model. GCG identifies global heads and target blocks, and strengthens global-context features directly within a single forward pass. Extensive experiments on various diffusion LLMs and benchmarks demonstrate that GCG consistently improves performance over existing guidance methods, while incurring much lower inference overhead.

\section*{Acknowledgements}
This work was supported by Institute of Information \& Communications Technology Planning \& Evaluation (IITP) grants funded by the Korea government (MSIT) (No. RS-2022-00143524, Development of Fundamental Technology and Integrated Solution for Next-Generation Automatic Artificial Intelligence System, and No. RS-2026-25519622).

{
\small
\bibliographystyle{unsrt}
\bibliography{references}

@article{nie2025large,
  title={Large Language Diffusion Models},
  author={Nie, Shen and Zhu, Fengqi and You, Zebin and Zhang, Xiaolu and Ou, Jingyang and Hu, Jun and Zhou, Jun and Lin, Yankai and Wen, Ji-Rong and Li, Chongxuan},
  journal={arXiv preprint arXiv:2502.09992},
  year={2025}
}

@article{ye2025dream,
  title={Dream 7B: Diffusion Large Language Models},
  author={Ye, Jiacheng and Xie, Zhihui and Zheng, Lin and Gao, Jiahui and Wu, Zirui and Jiang, Xin and Li, Zhenguo and Kong, Lingpeng},
  journal={arXiv preprint arXiv:2508.15487},
  year={2025}
}

@article{li2022diffusion,
  title={Diffusion-lm improves controllable text generation},
  author={Li, Xiang and Thickstun, John and Gulrajani, Ishaan and Liang, Percy S and Hashimoto, Tatsunori B},
  journal={NeurIPS},
  year={2022}
}

@article{gong2022diffuseq,
  title={Diffuseq: Sequence to sequence text generation with diffusion models},
  author={Gong, Shansan and Li, Mukai and Feng, Jiangtao and Wu, Zhiyong and Kong, LingPeng},
  journal={arXiv preprint arXiv:2210.08933},
  year={2022}
}

@article{nie2024scaling,
  title={Scaling up masked diffusion models on text},
  author={Nie, Shen and Zhu, Fengqi and Du, Chao and Pang, Tianyu and Liu, Qian and Zeng, Guangtao and Lin, Min and Li, Chongxuan},
  journal={arXiv preprint arXiv:2410.18514},
  year={2024}
}

@article{grattafiori2024llama,
  title={The llama 3 herd of models},
  author={Grattafiori, Aaron and Dubey, Abhimanyu and Jauhri, Abhinav and Pandey, Abhinav and Kadian, Abhishek and Al-Dahle, Ahmad and Letman, Aiesha and Mathur, Akhil and Schelten, Alan and Vaughan, Alex and others},
  journal={arXiv preprint arXiv:2407.21783},
  year={2024}
}

@article{yang2025qwen3,
  title={Qwen3 technical report},
  author={Yang, An and Li, Anfeng and Yang, Baosong and Zhang, Beichen and Hui, Binyuan and Zheng, Bo and Yu, Bowen and Gao, Chang and Huang, Chengen and Lv, Chenxu and others},
  journal={arXiv preprint arXiv:2505.09388},
  year={2025}
}

@article{liu2024deepseek,
  title={Deepseek-v3 technical report},
  author={Liu, Aixin and Feng, Bei and Xue, Bing and Wang, Bingxuan and Wu, Bochao and Lu, Chengda and Zhao, Chenggang and Deng, Chengqi and Zhang, Chenyu and Ruan, Chong and others},
  journal={arXiv preprint arXiv:2412.19437},
  year={2024}
}

@article{ho2022classifier,
  title={Classifier-free diffusion guidance},
  author={Ho, Jonathan and Salimans, Tim},
  journal={NeurIPSW},
  year={2021}
}

@article{li2025adaptive,
  title={Adaptive classifier-free guidance via dynamic low-confidence masking},
  author={Li, Pengxiang and Yan, Shilin and Tsai, Joey and Zhang, Renrui and An, Ruichuan and Guo, Ziyu and Gao, Xiaowei},
  journal={NeurIPS},
  year={2025}
}

@article{karras2024guiding,
  title={Guiding a diffusion model with a bad version of itself},
  author={Karras, Tero and Aittala, Miika and Kynk{\"a}{\"a}nniemi, Tuomas and Lehtinen, Jaakko and Aila, Timo and Laine, Samuli},
  journal={NeurIPS},
  year={2024}
}

@article{wu2025fast,
  title={Fast-dllm: Training-free acceleration of diffusion llm by enabling kv cache and parallel decoding},
  author={Wu, Chengyue and Zhang, Hao and Xue, Shuchen and Liu, Zhijian and Diao, Shizhe and Zhu, Ligeng and Luo, Ping and Han, Song and Xie, Enze},
  journal={ICLR},
  year={2026}
}

@article{li2025beyond,
  title={Beyond fixed: Training-free variable-length denoising for diffusion large language models},
  author={Li, Jinsong and Dong, Xiaoyi and Zang, Yuhang and Cao, Yuhang and Wang, Jiaqi and Lin, Dahua},
  journal={ICLR},
  year={2026}
}

@article{ma2025dkv,
  title={dkv-cache: The cache for diffusion language models},
  author={Ma, Xinyin and Yu, Runpeng and Fang, Gongfan and Wang, Xinchao},
  journal={NeurIPS},
  year={2025}
}

@article{wang2025remasking,
  title={Remasking discrete diffusion models with inference-time scaling},
  author={Wang, Guanghan and Schiff, Yair and Sahoo, Subham Sekhar and Kuleshov, Volodymyr},
  journal={NeurIPS},
  year={2025}
}

@article{hong2025wide,
  title={Wide-in, narrow-out: Revokable decoding for efficient and effective dllms},
  author={Hong, Feng and Yu, Geng and Ye, Yushi and Huang, Haicheng and Zheng, Huangjie and Zhang, Ya and Wang, Yanfeng and Yao, Jiangchao},
  journal={ICLR},
  year={2026}
}

@article{ma2026diffusion,
  title={Diffusion In Diffusion: Reclaiming Global Coherence in Semi-Autoregressive Diffusion},
  author={Ma, Linrui and Cui, Yufei and Han, Kai and Wang, Yunhe},
  journal={arXiv},
  year={2026}
}

@article{dhariwal2021diffusion,
  title={Diffusion models beat gans on image synthesis},
  author={Dhariwal, Prafulla and Nichol, Alexander},
  journal={NeurIPS},
  year={2021}
}

@article{clark2018arc,
  title={Think you have Solved Question Answering? Try ARC, the AI2 Reasoning Challenge},
  author={Clark, Peter and Cowhey, Isaac and Etzioni, Oren and Khot, Tushar and Sabharwal, Ashish and Schoenick, Carissa and Tafjord, Oyvind},
  year={2018},
  journal={arXiv}
}

@inproceedings{bisk2020piqa,
  title={PIQA: Reasoning about Physical Commonsense in Natural Language},
  author={Bisk, Yonatan and Zellers, Rowan and Gao, Jianfeng and Choi, Yejin and others},
  year={2020},
  booktitle={AAAI}
}

@article{chen2021humaneval,
  title={Evaluating Large Language Models Trained on Code},
  author={Chen, Mark and Tworek, Jerry and Jun, Heewoo and Yuan, Qiming and Pinto, Henrique Ponde De Oliveira and Kaplan, Jared and Edwards, Harri and Burda, Yuri and Joseph, Nicholas and Brockman, Greg and others},
  year={2021},
  journal={arXiv}
}

@article{austin2021mbpp,
  title={Program Synthesis with Large Language Models},
  author={Austin, Jacob and Odena, Augustus and Nye, Maxwell and Bosma, Maarten and Michalewski, Henryk and Dohan, David and Jiang, Ellen and Cai, Carrie and Terry, Michael and Le, Quoc and others},
  year={2021},
  journal={arXiv}
}

@article{cobbe2021gsm8k,
  title={Training Verifiers to Solve Math Word Problems},
  author={Cobbe, Karl and Kosaraju, Vineet and Bavarian, Mohammad and Chen, Mark and Jun, Heewoo and Kaiser, Lukasz and Plappert, Matthias and Tworek, Jerry and Hilton, Jacob and Nakano, Reiichiro and others},
  year={2021},
  journal={arXiv}
}

@inproceedings{hsieh2024ruler,
  title={RULER: What's the Real Context Size of Your Long-Context Language Models?},
   author={Hsieh, Cheng-Ping and Sun, Simeng and Kriman, Samuel and Acharya, Shantanu and Rekesh, Dima and Jia, Fei and Zhang, Yang and Ginsburg, Boris},
  year={2024},
  booktitle={COLM}
}

@inproceedings{huang2025minilongbench,
  title={MiniLongBench: The Low-cost Long Context Understanding Benchmark for Large Language Models},
  author={Huang, Zhongzhan and Ling, Guoming and Zhong, Shanshan and Wu, Hefeng and Lin, Liang},
  year={2025},
  booktitle={ACL}
}

@inproceedings{ahn2024self,
  title={Self-rectifying diffusion sampling with perturbed-attention guidance},
  author={Ahn, Donghoon and Cho, Hyoungwon and Min, Jaewon and Jang, Wooseok and Kim, Jungwoo and Kim, SeonHwa and Park, Hyun Hee and Jin, Kyong Hwan and Kim, Seungryong},
  booktitle={ECCV},
  year={2024},
}

@article{chen2025s,
  title={S $^2$-Guidance: Stochastic Self Guidance for Training-Free Enhancement of Diffusion Models},
  author={Chen, Chubin and Zhu, Jiashu and Feng, Xiaokun and Huang, Nisha and Wu, Meiqi and Mao, Fangyuan and Wu, Jiahong and Chu, Xiangxiang and Li, Xiu},
  journal={ICLR},
  year={2026}
}

@inproceedings{bai2024longbench,
  title={Longbench: A bilingual, multitask benchmark for long context understanding},
  author={Bai, Yushi and Lv, Xin and Zhang, Jiajie and Lyu, Hongchang and Tang, Jiankai and Huang, Zhidian and Du, Zhengxiao and Liu, Xiao and Zeng, Aohan and Hou, Lei and others},
  booktitle={ACL},
  year={2024}
}

@article{dao2022flashattention,
  title={Flashattention: Fast and memory-efficient exact attention with io-awareness},
  author={Dao, Tri and Fu, Dan and Ermon, Stefano and Rudra, Atri and R{\'e}, Christopher},
  journal={NeurIPS},
  year={2022}
}

@article{zhu2025llada,
  title={Llada 1.5: Variance-reduced preference optimization for large language diffusion models},
  author={Zhu, Fengqi and Wang, Rongzhen and Nie, Shen and Zhang, Xiaolu and Wu, Chunwei and Hu, Jun and Zhou, Jun and Chen, Jianfei and Lin, Yankai and Wen, Ji-Rong and others},
  journal={arXiv preprint arXiv:2505.19223},
  year={2025}
}

@book{golub2013matrix,
  title={Matrix computations},
  author={Golub, Gene H and Van Loan, Charles F},
  year={2013},
  publisher={JHU press}
}

@article{liu2019roberta,
  title={Roberta: A robustly optimized bert pretraining approach},
  author={Liu, Yinhan and Ott, Myle and Goyal, Naman and Du, Jingfei and Joshi, Mandar and Chen, Danqi and Levy, Omer and Lewis, Mike and Zettlemoyer, Luke and Stoyanov, Veselin},
  journal={arXiv preprint arXiv:1907.11692},
  year={2019}
}

@article{wang2019glue,
  title={GLUE: A multi-task benchmark and analysis platform for natural language understanding},
  author={Wang, Alex and Singh, Amanpreet and Michael, Julian and Hill, Felix and Levy, Omer and Bowman, Samuel R},
  journal={ICLR},
  year={2019}
}

@inproceedings{rajpurkar2018squad,
  title={Know what you don’t know: Unanswerable questions for SQuAD},
  author={Rajpurkar, Pranav and Jia, Robin and Liang, Percy},
  booktitle={Proceedings of the 56th Annual Meeting of the Association for Computational Linguistics (Volume 2: Short Papers)},
  year={2018}
}
}

\clearpage

\appendix

\section{Synthetic infilling benchmark}
\label{sec:app_synthetic}

We provide additional details on the synthetic infilling benchmark used in our analysis. Each sample is generated from predefined pools of synthetic entities, including 24 clients, 24 projects, 24 project codenames, 24 deployment channels, and 768 approval owners. The owner pool is constructed by combining 32 first-name identifiers with 24 last-name identifiers. Each pool contains synthetic names such as Orion, Nova, Vega, and Altair for clients; Zephyr, Meridian, Aurora, and Nimbus for projects; Aster, Northwind, Bluehaven, and Ironleaf for codenames; and North-7, East-2, West-4, and South-9 for deployment channels. We generate total 700 samples.

For each sample, we randomly sample a subset of entities and define mappings from clients to projects and from projects to their attributes. We then choose a target client and construct a document with masked tokens in the middle. The masked tokens correspond to the project codename, deployment channel, and approval owner of the project assigned to the target client. Solving the task therefore requires the model to first identify the project associated with the target client and then retrieve the attributes of that project.

We use a bidirectional evidence structure. The client-to-project evidence is placed before the masked tokens, while the project-attribute evidence is placed after them. The document also includes filler text and distractor statements involving non-target clients and projects, so that the answer cannot be inferred from the surrounding template or the target client alone.

To study the effect of contextual distance, we randomize the amount of intervening context between the masked tokens and the supporting evidence. By mixing evidence sentences with filler text and distractor statements, the generated examples cover a broad range of realized evidence--mask distances, from approximately 5 to 300 tokens. This design allows us to compare local-context cases, where the evidence appears near the masked tokens, with global-context cases, where the evidence is separated by substantially longer context.

\begin{figure*}[!t]
  \small
  \centering

  \begin{subfigure}[t]{0.38\textwidth}
    \centering
    \includegraphics[height=3.2cm]{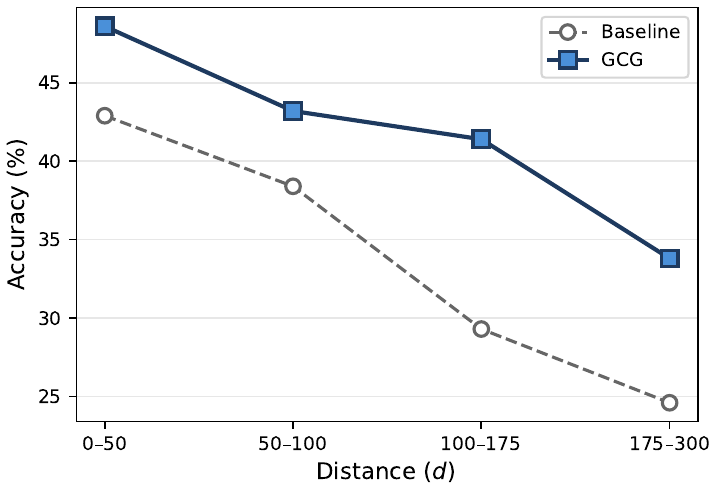}
    \caption{Accuracy vs.\ evidence--mask distance.}
    \label{fig:dream_distance}
  \end{subfigure}
  \hfill
  \begin{subfigure}[t]{0.58\textwidth}
    \centering
    \includegraphics[height=3.2cm]{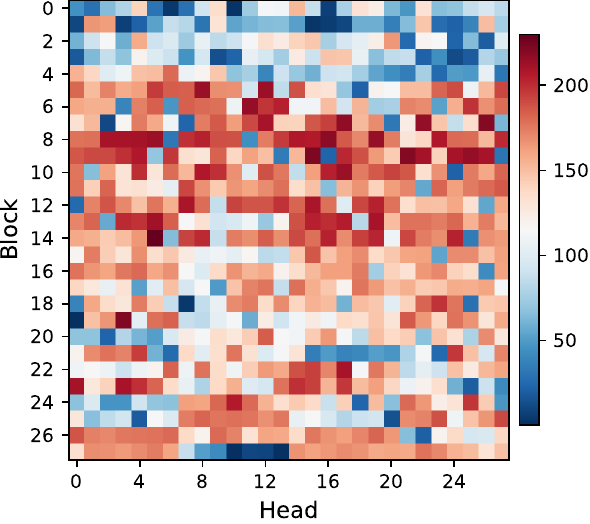}
    \caption{COM of attention heads across blocks.}
    \label{fig:dream_com_heatmap}
  \end{subfigure}

  \vspace{0.5em}

  \begin{subfigure}[t]{0.38\textwidth}
    \centering
    \includegraphics[height=2.8cm,width=\linewidth,keepaspectratio]{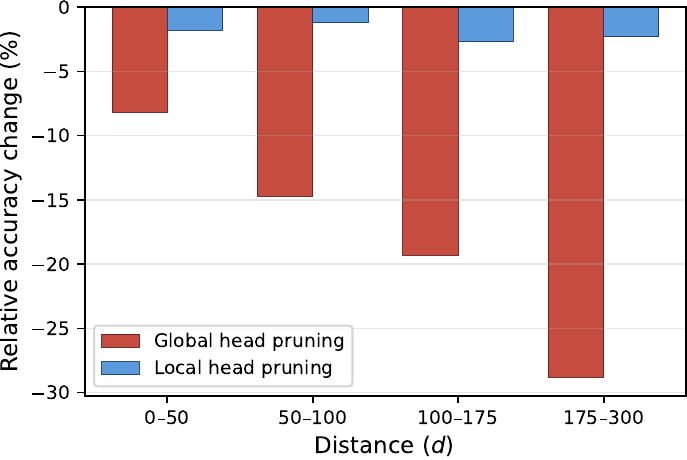}
    \caption{Global vs.\ local head pruning.}
    \label{fig:dream_prune_com}
  \end{subfigure}
  \hfill
  \begin{subfigure}[t]{0.58\textwidth}
    \centering
    \includegraphics[height=2.8cm,width=\linewidth,keepaspectratio]{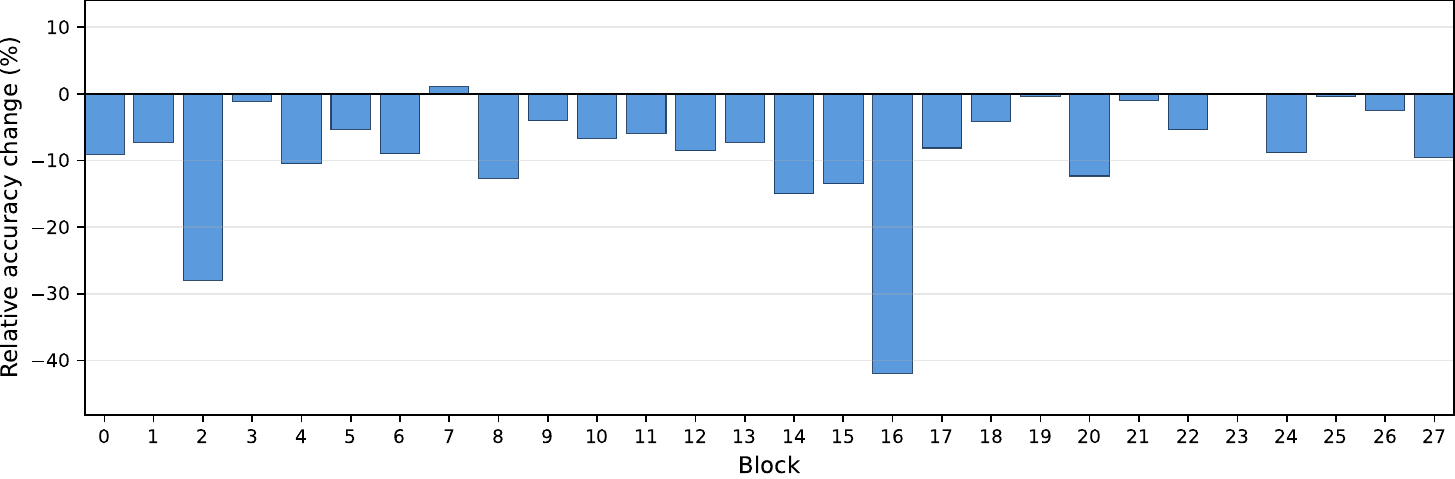}
    \caption{Block-wise pruning of global heads.}
    \label{fig:dream_layer_knockout}
  \end{subfigure}

  \caption{
Additional analyses of Dream-7B~\cite{ye2025dream} using the synthetic infilling benchmark. (a) Accuracy as a function of the average evidence--mask distance. (b) COM heatmap across all blocks and heads. (c) Relative accuracy change after pruning global heads (red) or local heads (blue) across evidence--mask distances. (d) Block-wise relative accuracy change after removing global heads from each block individually.
  }
  \label{fig:dream_analysis_appendix}
\end{figure*}

\begin{figure*}[!t]
  \small
  \centering

  \begin{subfigure}[t]{0.38\textwidth}
    \centering
    \includegraphics[height=3.2cm]{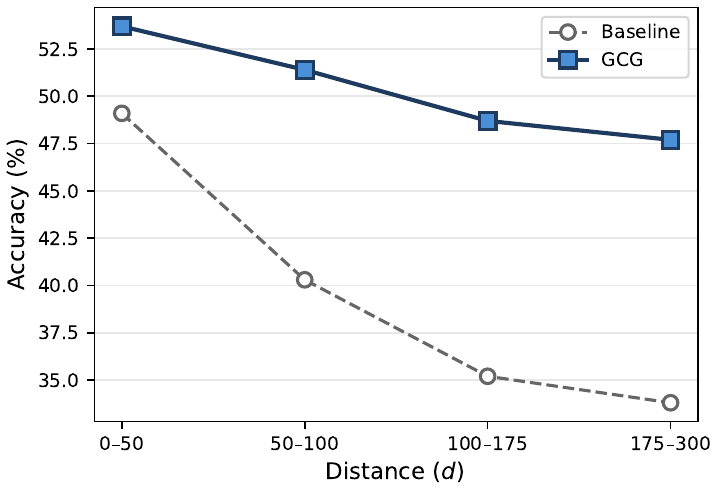}
    \caption{Accuracy vs.\ evidence--mask distance.}
    \label{fig:llada15_distance}
  \end{subfigure}
  \hfill
  \begin{subfigure}[t]{0.58\textwidth}
    \centering
    \includegraphics[height=3.2cm]{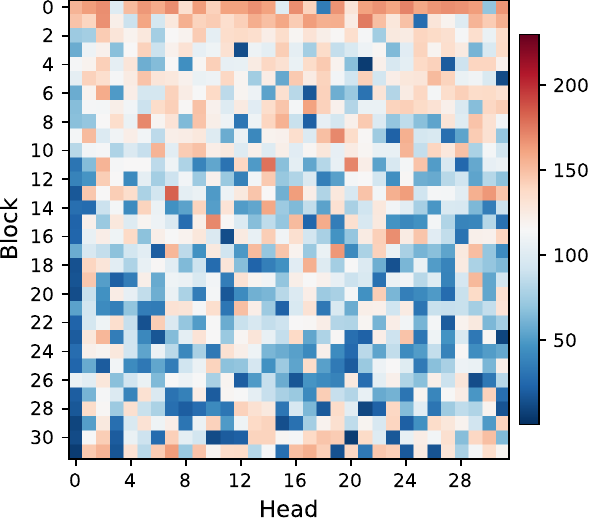}
    \caption{COM of attention heads across blocks.}
    \label{fig:llada15_com_heatmap}
  \end{subfigure}

  \vspace{0.5em}

  \begin{subfigure}[t]{0.38\textwidth}
    \centering
    \includegraphics[height=2.8cm,width=\linewidth,keepaspectratio]{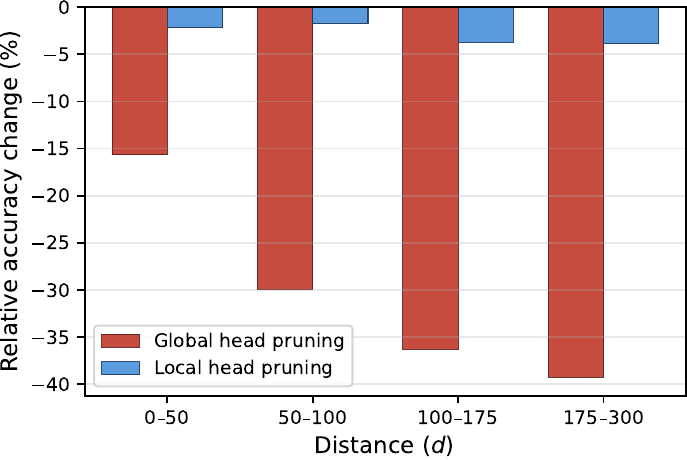}
    \caption{Global vs.\ local head pruning.}
    \label{fig:llada15_prune_com}
  \end{subfigure}
  \hfill
  \begin{subfigure}[t]{0.58\textwidth}
    \centering
    \includegraphics[height=2.8cm,width=\linewidth,keepaspectratio]{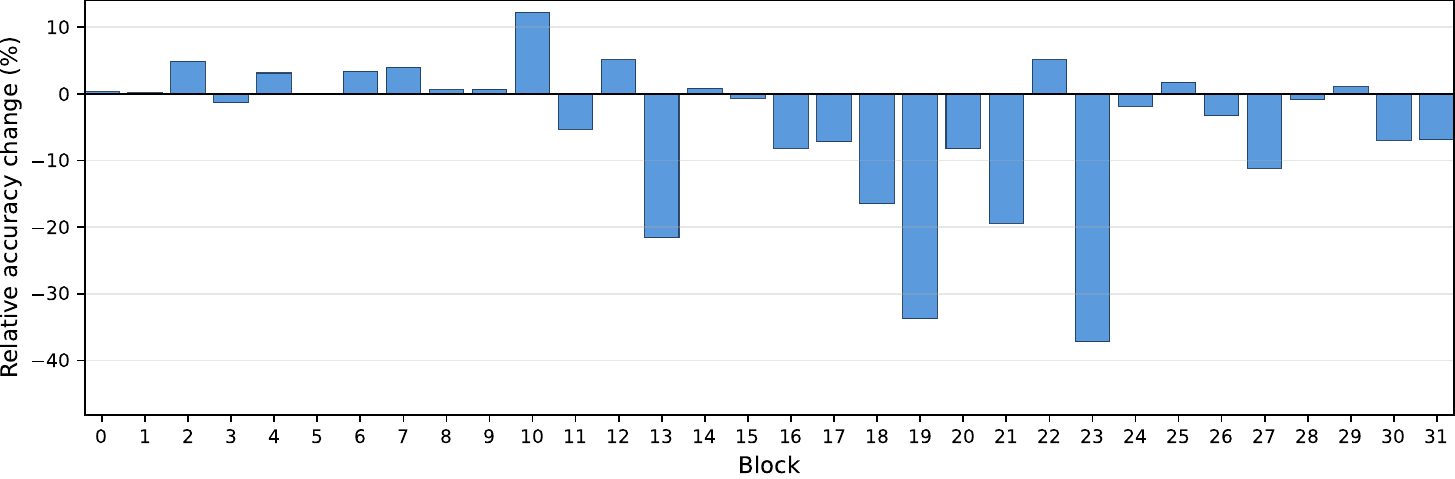}
    \caption{Block-wise pruning of global heads.}
    \label{fig:llada15_layer_knockout}
  \end{subfigure}

  \caption{
Additional analyses of LLaDA-1.5~\cite{zhu2025llada} using the synthetic infilling benchmark. (a) Accuracy as a function of the average evidence--mask distance. (b) COM heatmap across all blocks and heads. (c) Relative accuracy change after pruning global heads (red) or local heads (blue) across evidence--mask distances. (d) Block-wise relative accuracy change after removing global heads from each block individually.
  }
  \label{fig:llada15_analysis_appendix}
\end{figure*}

\section{More implementation details}
\label{sec:app_imp}

\paragraph{Calibration procedure.}
We describe the procedure for selecting global heads, local heads, and target blocks in Sec.~\ref{sec:GCG}.
Note that the synthetic infilling benchmark in Sec.~\ref{sec:analysis} is used only to analyze the use of contextual information under controlled evidence--mask distances, and it is not used for the selection.
Instead, we use natural examples from LongBench~\cite{bai2024longbench}.
Specifically, we randomly sample three examples from each of the 21 tasks in LongBench, resulting in 63 calibration examples that cover question answering, summarization, code, and dialogue.
The selection process consists of three steps:
(1) We run the standard denoising process on each calibration example, and compute the COM score of each head $h$ in block $m$ using Eq.~(3).
We then average the COM scores over the calibration examples.
(2) We rank the heads in each block by their average COM scores, and select the top-$p\%$ and bottom-$p\%$ heads as global heads $\mathcal{G}^m$ and local heads $\mathcal{L}^m$, respectively.
(3) We prune the global heads in a single block by zeroing their outputs, and measure the resulting performance drop on the calibration examples.
We repeat this for each block, and select the top-$q\%$ blocks with the largest performance drops as the set of target blocks $\mathcal{M}$.
We set both $p$ and $q$ to 25 by default.
This process is performed once for each model.
The selected $\mathcal{G}^m$, $\mathcal{L}^m$, and $\mathcal{M}$ are then fixed across all evaluation benchmarks.

\paragraph{Decoding procedure.}
We provide additional details on the evaluation settings used in our experiments. For multiple choice benchmarks, including ARC~\cite{clark2018arc} and PIQA~\cite{bisk2020piqa}, we follow the common evaluation protocol for diffusion LLMs, scoring each answer candidate using Monte Carlo likelihood estimation. Specifically, given a prompt and a candidate answer, we repeatedly sample a subset of answer tokens, replace them with masked tokens, and compute the log probability assigned to the original tokens at the masked positions. Each estimate is reweighted by the ratio between the answer length and the number of masked tokens. We average the resulting scores over 128 Monte Carlo samples and select the answer candidate with the highest estimated likelihood.

For generative benchmarks~\cite{chen2021humaneval, austin2021mbpp, cobbe2021gsm8k, hsieh2024ruler, huang2025minilongbench}, we set the generation length and denoising steps according to the expected output length of each task. For HumanEval~\cite{chen2021humaneval} and MBPP~\cite{austin2021mbpp}, we generate up to 256 tokens with 64 denoising steps. For GSM8K-CoT~\cite{cobbe2021gsm8k}, we generate up to 512 tokens with 128 denoising steps. We set the number of few-shot examples to 0, 3, and 8 for HumanEval~\cite{chen2021humaneval}, MBPP~\cite{austin2021mbpp}, and GSM8K-CoT~\cite{cobbe2021gsm8k}, respectively. For long context benchmarks, we set the prompt length to 8K tokens for both Ruler and MiniLongBench. We use 64 generation tokens and 16 denoising steps for Ruler, and 512 generation tokens and 128 denoising steps for MiniLongBench.

We follow the official implementation of each model for decoding. For LLaDA~\cite{nie2025large} and LLaDA-1.5~\cite{zhu2025llada}, we unmask tokens in descending order of prediction confidence, so that positions with higher confidence are filled earlier during denoising. For Dream-7B~\cite{ye2025dream}, we follow its default random unmasking strategy, where token positions are selected randomly at each denoising step.

\begin{table}[t]
\centering
\caption{Quantitative analysis of GCG with different head and block selection percentages on Dream-7B~\cite{ye2025dream}. The parameters $p$ and $q$ control the percentages of selected heads and target blocks, respectively. Best scores in each column are in bold.}
\label{tab:hyperparam_pq_dream7b}
\resizebox{0.95\linewidth}{!}{
\begin{tabular}{ccccccccc}
\toprule
$p$ & $q$ & ARC & PIQA & HumanEval & MBPP & GSM8K-CoT & Ruler & MiniLongBench \\
\midrule
0.0 & 0.0 & 54.2 & 75.5 & 1.8 & 15.0 & 3.9 & 28.0 & 26.5 \\
\midrule
\multirow{3}{*}{12.5}
& 12.5 & 58.4 & 75.7 & 1.8 & 15.0 & 5.1 & 29.8 & 26.8 \\
& 25.0 & 58.3 & 75.5 & 3.0 & 15.8 & 4.5 & 28.0 & 26.5 \\
& 50.0 & 58.8 & 75.7 & 3.0 & 15.4 & 5.2 & 30.8 & \textbf{27.9} \\
\midrule
\multirow{3}{*}{25.0}
& 12.5 & 58.8 & 75.8 & 1.8 & 15.8 & 4.9 & 29.5 & 26.8 \\
& 25.0 & {58.4} & {75.6} & 3.1 & {\textbf{17.0}} & {5.4} & {28.5} & {27.1} \\
& 50.0 & 58.9 & 75.8 & 3.7 & 14.4 & 4.2 & 30.4 & 26.6 \\
\midrule
\multirow{3}{*}{50.0}
& 12.5 & 58.5 & 75.6 & 3.7 & 15.0 & 5.2 & 29.4 & 26.6 \\
& 25.0 & 58.4 & 75.5 & \textbf{4.3} & 16.0 & \textbf{6.2} & 28.5 & 26.9 \\
& 50.0 & \textbf{59.0} & \textbf{75.9} & 1.8 & 16.6 & 5.4 & \textbf{32.2} & 26.7 \\
\bottomrule
\end{tabular}
}
\end{table}

\begin{table}[!th]
\centering
\caption{Quantitative analysis of GCG with different head and block selection percentages on LLaDA-1.5~\cite{zhu2025llada}. The parameters $p$ and $q$ control the percentages of selected heads and target blocks, respectively. Best scores in each column are in bold.}
\label{tab:hyperparam_pq_llada15}
\resizebox{0.95\linewidth}{!}{
\begin{tabular}{ccccccccc}
\toprule
$p$ & $q$ & ARC & PIQA & HumanEval & MBPP & GSM8K-CoT & Ruler & MiniLongBench \\
\midrule
0.0 & 0.0 & 51.4 & 73.5 & 15.9 & 17.2 & 37.8 & 37.6 & 31.6 \\
\midrule
\multirow{3}{*}{12.5}
& 12.5 & 54.8 & 73.6 & 17.1 & 18.8 & 37.1 & 37.4 & \textbf{32.8} \\
& 25.0 & 54.7 & 73.9 & 17.1 & 18.8 & 38.9 & 37.6 & 32.2 \\
& 50.0 & \textbf{55.2} & 74.0 & 18.3 & 18.2 & 38.5 & 38.8 & 32.6 \\
\midrule
\multirow{3}{*}{25.0}
& 12.5 & 54.4 & 73.9 & 17.7 & 19.0 & 39.4 & 37.9 & 32.5 \\
& 25.0 & {54.5} & {73.9} & {\textbf{18.9}} & 19.0 & {38.5} & {39.3} & {32.2} \\
& 50.0 & 54.8 & \textbf{74.1} & 18.3 & 18.0 & \textbf{40.9} & 37.4 & 32.7 \\
\midrule
\multirow{3}{*}{50.0}
& 12.5 & 54.6 & 73.8 & 18.3 & 17.8 & 37.5 & 41.6 & 32.0 \\
& 25.0 & 54.6 & 73.9 & 17.1 & 18.0 & 38.8 & \textbf{42.3} & 32.3 \\
& 50.0 & 55.1 & 73.9 & \textbf{18.9} & \textbf{19.8} & 39.3 & 39.7 & 31.9 \\
\bottomrule
\end{tabular}
}
\end{table}

\section{Additional analyses}
\label{sec:app_results}

We provide additional analyses on Dream-7B~\cite{ye2025dream} and LLaDA-1.5~\cite{zhu2025llada}.

\paragraph{Analysis on context utilization.}
We provide additional analyses on context utilization for Dream-7B~\cite{ye2025dream} and LLaDA-1.5~\cite{zhu2025llada} in Figs.~\ref{fig:dream_analysis_appendix} and~\ref{fig:llada15_analysis_appendix}, respectively. We follow the same analysis protocol as in Sec.~\ref{sec:analysis} using the synthetic infilling benchmark. Similar to the results on LLaDA-8B, both models show a consistent drop in accuracy as the average evidence--mask distance increases, indicating that they rely more heavily on local context and underutilize global context during denoising. The COM heatmaps also show that attention heads exhibit clear variation in their average attention distances, suggesting the existence of heads that attend more strongly to distant positions. We further observe that pruning global heads leads to larger performance degradation than pruning local heads, especially when the supporting evidence is distant. Finally, the block-wise pruning results show that the effect of removing global heads varies substantially across blocks. These results confirm that the main observations in Sec.~\ref{sec:analysis} hold across different diffusion LLMs, where  diffusion LLMs are biased toward local context, and a subset of global heads and blocks plays an important role in leveraging global context.

\paragraph{Analysis on hyperparameters.}
We further evaluate the sensitivity of GCG to the head and block selection ratios, $p$ and $q$, and the guidance scale $s$ on Dream-7B and LLaDA-1.5. Tables~\ref{tab:hyperparam_pq_dream7b} and~\ref{tab:hyperparam_pq_llada15} show the results obtained by varying $p$ and $q$, while Tables~\ref{tab:hyperparam_s_dream7b} and~\ref{tab:hyperparam_s_llada15} show the results obtained by varying $s$. We can see that GCG generally improves the baseline across a wide range of settings for both models. These results suggest that GCG is not highly sensitive to a specific hyperparameter configuration, and that its gains do not rely on careful hyperparameter tuning.

\begin{table}[t]
\centering
\caption{Quantitative analysis of GCG with different guidance scales on Dream-7B~\cite{ye2025dream}. The setting $s=0$ corresponds to the baseline without guidance. Best scores in each column are in bold.}
\label{tab:hyperparam_s_dream7b}
\resizebox{0.85\linewidth}{!}{
\begin{tabular}{cccccccc}
\toprule
$s$ & ARC & PIQA & HumanEval & MBPP & GSM8K-CoT & Ruler & MiniLongBench \\
\midrule
0.0 & 54.2 & 75.5 & 1.8 & 15.0 & 3.9 & 28.0 & 26.5\\
0.1 & 58.5 & 75.5 & 2.4 & \textbf{17.6} & 5.2 & 28.7 & \textbf{27.4} \\
0.2 & 58.3 & 75.6 & 2.4 & 15.8 & 5.1 & \textbf{28.9} & \textbf{27.4} \\
0.3 & {58.4} & {75.6} & 3.1 & {17.0} & {5.4} & {28.5} & {27.1} \\
0.4 & 58.4 & 75.6 & 3.1 & 14.6 & \textbf{5.5} & 28.1 & 26.5 \\
0.5 & \textbf{58.6} & \textbf{76.4} & \textbf{4.3} & 16.6 & \textbf{5.5} & 28.1 & 26.9 \\
\bottomrule
\end{tabular}
}
\end{table}

\begin{table}[t]
\centering
\caption{Quantitative analysis of GCG with different guidance scales on LLaDA-1.5~\cite{zhu2025llada}. The setting $s=0$ corresponds to the baseline without guidance. Best scores in each column are in bold.}
\label{tab:hyperparam_s_llada15}
\resizebox{0.85\linewidth}{!}{
\begin{tabular}{cccccccc}
\toprule
$s$ & ARC & PIQA & HumanEval & MBPP & GSM8K-CoT & Ruler & MiniLongBench \\
\midrule
0.0 & 51.4 & 73.5 & 15.9 & 17.2 & 37.8 & 37.6 & 31.6 \\
0.1 & 54.4 & 73.8 & 17.1 & 18.6 & 38.4 & 38.0 & 31.8 \\
0.2 & 54.3 & 73.8 & \textbf{18.9} & 18.8 & \textbf{39.0} & 39.3 & 32.5 \\
0.3 & {54.5} & {\textbf{73.9}} & {\textbf{18.9}} & {\textbf{19.0}} & {38.5} & {39.3} & {32.2} \\
0.4 & \textbf{54.6} & \textbf{73.9} & \textbf{18.9} & 16.8 & 38.4 & \textbf{39.5} & 32.6 \\
0.5 & \textbf{54.6} & 73.8 & 18.3 & \textbf{19.0} & \textbf{39.0} & 38.5 & \textbf{33.5} \\
\bottomrule
\end{tabular}
}
\end{table}

\section{Broader impacts}
\label{sec:app_broader_impacts}

This work aims to improve diffusion large language models (LLMs) by making guidance more effective and efficient. A potential positive impact is that GCG reduces the inference overhead of guided generation by avoiding the additional forward pass required by existing guidance methods. This can make diffusion LLMs more computationally efficient, potentially reducing energy consumption and improving accessibility in resource-constrained settings.

However, improvements in language generation may also have negative societal impacts. More efficient and capable LLMs could be misused to generate misinformation, spam, phishing messages, or other harmful content. Moreover, since GCG operates on existing diffusion LLMs, it may inherit biases, privacy risks, hallucinations, and other limitations of the underlying models. Although our work is foundational and does not deploy a user-facing system, we encourage appropriate safeguards when applying GCG, such as safety evaluation, content moderation, abuse monitoring, and human oversight in high-stakes applications.

\end{document}